\documentclass[letterpaper, 10 pt, conference]{ieeeconf}  

\IEEEoverridecommandlockouts                              

\usepackage{colortbl}
\usepackage{xcolor}
\usepackage[utf8]{inputenc} 
\usepackage[T1]{fontenc}    
\usepackage{hyperref}       
\usepackage{url}            
\usepackage{booktabs}       
\usepackage{amsfonts}       
\usepackage{nicefrac}       
\usepackage{microtype}      
\usepackage{xcolor}         
\usepackage{graphicx}
\usepackage{multirow}
\usepackage{bm}
\usepackage{pifont}
\usepackage{subfigure}
\usepackage{amsmath}
\usepackage{subfiles}
\usepackage{hyperref}       

\title{\LARGE \bf
FG-Depth: Flow-Guided Unsupervised Monocular Depth Estimation
}

\author{Junyu Zhu$^{1}$, Lina Liu$^{1*}$, Yong Liu$^{1*}$, Wanlong Li$^{2}$, Feng Wen$^{2}$ and Hongbo Zhang$^{2}$
\thanks{$^{1}$Junyu Zhu, Lina Liu and Yong Liu are with the Institute of Cyber-Systems and Control, Zhejiang University, 
Hangzhou, China. email:\{junyuzhu, linaliu\}@zju.edu.cn, yongliu@iipc.zju.edu.cn. }
\thanks{$^{2}$Wanlong Li, Feng Wen and Hongbo Zhang are with Noah’s Ark Lab, Huawei Technologies, Beijing, China. email:\{liwanlong, wenfeng3, zhanghongbo888\}@huawei.com.}
\thanks{$^{*}$Corresponding authors: Lina Liu and Yong Liu.}
}%

\begin{document}

\maketitle
\thispagestyle{empty}
\pagestyle{empty}

\begin{abstract}

The great potential of unsupervised monocular depth estimation has been demonstrated by many works due to low annotation cost and impressive accuracy comparable to supervised methods. To further improve the performance, recent works mainly focus on designing more complex network structures and exploiting extra supervised information, e.g., semantic segmentation. 
These methods optimize the models by exploiting the reconstructed relationship between the target and reference images in varying degrees. However, previous methods prove that this image reconstruction optimization is prone to get trapped in local minima. 
In this paper, our core idea is to guide the optimization with prior knowledge from pretrained Flow-Net. And we show that the bottleneck of unsupervised monocular depth estimation can be broken with our simple but effective framework named FG-Depth. 
In particular, we propose (i) a flow distillation loss to replace the typical photometric loss that limits the capacity of the model and (ii) a prior flow based mask to remove invalid pixels that bring the noise in training loss.
Extensive experiments demonstrate the effectiveness of each component, and our approach achieves state-of-the-art results on both KITTI and NYU-Depth-v2 datasets.

\end{abstract}

\section{Introduction}

Accurate depth estimation is critical for many applications in computer vision, such as robotic perception~\cite{JFR2009,michels2005high}, augmented reality~\cite{karsch2014automatic}, and 3D modeling~\cite{2008Make3D}. Monocular depth estimation has become a challenging and promising field, attracting the attention of many researchers. Recently, deep learning-based monocular depth estimation~\cite{DORN,BTS,LapDepth,AdaBins,GLP_Depth} has been able to achieve high accuracy by narrowing the gap between predicted depth and ground truth. However, these methods are limited by the expensive annotation cost. The emergence of self-supervised approaches~\cite{Monodepth,Monodepth2,SfMLearner} addresses problems requiring depth annotations, typically trained using the photometric loss to reconstruct and warp images between target frames and source frames from monocular videos, stereo pairs, or stereo videos. The photometric loss widely used in self-supervised depth estimation is based on implicit assumptions~\cite{Monodepth} that 1) the scene is static; 2) no occlusion occurs between target frames and source frames; 3) the surface is Lambertian. However, these assumptions are so hard to be met on real data that the optimization of photometric loss is prone to be trapped in local minima~\cite{Depth_Hints} and the performance of the model is limited. 

\begin{figure}[h]
    \centering
    \begin{minipage}[t]{0.3\linewidth}
        \centerline{\includegraphics[width=\textwidth]{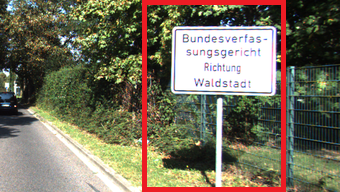}}
        \vspace{3pt}
        \centerline{\includegraphics[width=\textwidth]{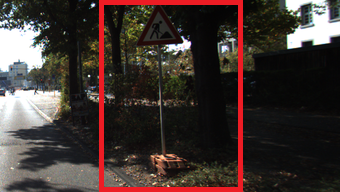}}
        \centerline{Input}
    \end{minipage}
    \begin{minipage}[t]{0.124\linewidth}
        \centerline{\includegraphics[width=\textwidth]{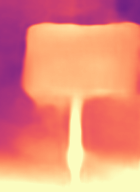}}
        \vspace{3pt}
        \centerline{\includegraphics[width=\textwidth]{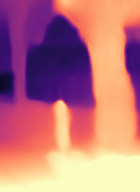}}
        \centerline{(a)}
    \end{minipage}
    \begin{minipage}[t]{0.124\linewidth}
        \centerline{\includegraphics[width=\textwidth]{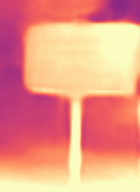}}
        \vspace{3pt}
        \centerline{\includegraphics[width=\textwidth]{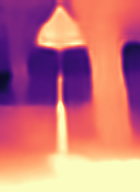}}
        \centerline{(b)}
    \end{minipage}
    \begin{minipage}[t]{0.124\linewidth}
        \centerline{\includegraphics[width=\textwidth]{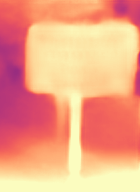}}
        \vspace{3pt}
        \centerline{\includegraphics[width=\textwidth]{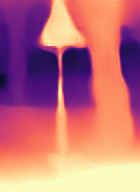}}
        \centerline{(c)}
    \end{minipage}
    \begin{minipage}[t]{0.124\linewidth}
        \centerline{\includegraphics[width=\textwidth]{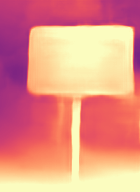}}
        \vspace{3pt}
        \centerline{\includegraphics[width=\textwidth]{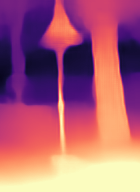}}
        \centerline{(d)}
    \end{minipage}
    \caption{\textbf{Comparison with state-of-the-art methods.} (a)MonoDepth2~\cite{Monodepth2}, (b)DepthHints~\cite{Depth_Hints}, (c)EPCDepth~\cite{EPCDepth}, (d)Ours.}\label{SIDE_results}
\end{figure}

Regarding the above problems, Waston et al.~\cite{Depth_Hints} used SGM algorithm results as depth hints to guide the model to reach better minima. Zhan et al.~\cite{Depth_VO_Feat} and Shu et al.~\cite{Feat_Depth} proposed the feature reconstruction loss to make the training loss more sensitive to low-texture regions and more robust to illumination change. \cite{Monodepth} uses stereo pair based photometric loss to avoid the influence of dynamic objects in image warping. Other methods ignore the defects of photometric loss and try to improve depth estimation performance by introducing additional semantic segmentation constraints, such as \cite{Edge_Depth}. They use semantic segmentation constraints to further the depth quality near object boundaries. However, annotating semantic segmentation in real data is expensive. Although low-cost semantic segmentation labels can be easily obtained from synthetic data, existing semantic segmentation models trained on synthetic data cannot generalize well to real data due to the domain shift~\cite{S3Net}. 

Despite many improvements, almost all existing unsupervised methods rely heavily on photometric loss with non-negligible defects, and performance has reached a bottleneck. Therefore, we believe that it is hard to make significant progress if the training process still relies on typical photometric loss.

In order to break the bottleneck of unsupervised monocular depth estimation, inspired by \cite{Guo_2018_ECCV}, in this paper, we design a new loss to replace the widely used photometric loss. 
Similar to \cite{Guo_2018_ECCV}, our Depth-Net also learns monocular depth by distilling prior knowledge of an optical flow estimation network that has strong generalization and can still generalize well to real data when trained on low-cost synthetic data. But there are three main differences between our method and \cite{Guo_2018_ECCV}. Besides the intuitive depth level, our proposed loss further restricts the Depth-Net from the color level. And we propose a mask to filter out out-of-range pixels at distillation time to accelerate convergence. Also, our network architectures are different from theirs. 

Our framework is trained based on stereo pairs to avoid influence from moving objects. 
Firstly, based on the fact that the depth pseudo labels can be generated from prior flow for stereo pairs and the warping procedure in computing photometric loss is based on flow which can be synthesized from depth estimation or directly predicted by pretrained optical flow estimation network, we propose a flow distillation loss to restrict model from depth and color levels. 
Secondly, Depth-Net is usually only effective within a certain range, and pixels outside the estimated range also inhibit training because they always fail to match the corresponding pixels in the warping procedure. In the previous method~\cite{Monodepth2,HR_Depth}, there are some pixels beyond the above range that cannot be removed by the previous methods during the training. To this end, taking full advantage of the prior flow, we propose a mask to remove pixels outside the estimation range. 
Fig.~\ref{SIDE_results} shows that our approach (Fig.~\ref{SIDE_results}(d)) can break the bottleneck of self-supervised monocular depth estimation compared to other methods(Fig.~\ref{SIDE_results}(a) to (c)).

To summarize, our main contributions are:
\begin{itemize}
  \item We introduce a flow distillation loss for restricting the model from depth and color levels to replace the typical photometric loss.
  \item We propose a prior flow based mask for pixels out of the estimation range at distillation time to improve performance.
  \item The proposed model achieves state-of-the-art performance on the KITTI and NYU-Depth-v2 datasets.
\end{itemize}

\section{Related works}

\subsection{Supervised Monocular Depth Estimation}
For the task of supervised monocular depth estimation, ground truth depth labels are used to supervise the training of the model with RGB monocular images as input. The ground truth depth labels are usually captured with lidars or RGBD cameras which have the disadvantage of high cost or limited depth range and usage scenario. 
Most supervised methods regard monocular depth estimation as a regression task. 
Eigen et al.~\cite{Eigen} is the first to employ CNN in supervised monocular depth estimation. 
Xu et al.~\cite{2018CRF} applied CRF to optimally combine multi-scale information derived from the inner layers of CNN to improve the performance of a CNN depth estimator. 
Fu et al.~\cite{DORN} found that a performance improvement can be achieved when the depth estimation is regarded as a classification task. 
With more complex and well-designed CNN architectures, \cite{BTS,LapDepth} refreshed previous records. 
And in recent years, thanks to the development of ViT~\cite{ViT}, several models~\cite{GPT,AdaBins,GLP_Depth,NeWCRFs} have been proposed to help the accuracy reach new heights. 

\subsection{Unsupervised Monocular Depth Estimation}
To avoid the expensive cost of depth annotation, unsupervised methods usually use photometric loss between adjacent frames to train the model. 
As the earliest works in the self-supervised depth estimation field, \cite{Monodepth} uses stereo pairs to train the Depth-Net and \cite{SfMLearner} is trained by monocular video with an extra Pose-Net to predict the relative pose between adjacent frames. \cite{SfMLearner} introduced a mask predicted by the network to reduce the influence of occlusions and moving objects. Yin et al.~\cite{Geonet} and Wang et al.~\cite{UnOS} excavated more geometric constraints by learning the depth and flow jointly in an unsupervised manner.
Godard et al.~\cite{Monodepth2} made a noticeable improvement by proposing a minimum photometric loss to handle occlusions, an auto mask to ignore pixels that violate camera motion assumptions, and a full-resolution multi-scale sampling method to make the prediction more accurate. 
Noting that the model can struggle during the training to find the global optimum when minimizing the photometric loss because of low-texture regions and illumination change, \cite{Depth_Hints} introduced SGM algorithm results as extra constraints in training loss and \cite{Depth_VO_Feat,Feat_Depth} took reconstruction on feature level into consideration. 
To further improve the performance, some works~\cite{klingner2020self,Edge_Depth,S3Net} brought extra semantic segmentation constraints into the training loss but extra semantic labels on real data actually increase the burden of annotation. And recently, Peng et al.~\cite{EPCDepth} introduce an effective data augmentation method for stereo-based models. 

\subsection{Optical Flow Estimation}
Optical flow estimation is the task of estimating per-pixel displacement between two frames. Recently, many deep learning based approaches~\cite{FlowNet,FlowNet2,LiteFlowNet,PWCNet,MaskFlownet} have been proposed for optical flow estimation. 
In supervised optical flow estimation tasks, models are usually trained on synthetic data that has dense accurate optical flow labels. 
After the training on the synthetic data, they can usually generalize well on the real data. 
As an exceptional case of optical flow estimation, stereo matching has additional constraints that the displacement is always 
negative along the horizontal direction and always zero along the vertical direction. 
Also, many stereo matching models trained on synthetic data can generalize well on real data.

\section{Method}

\subsection{Method Overview}

\begin{figure*}[h]
    \centering
    \includegraphics[scale=0.5]{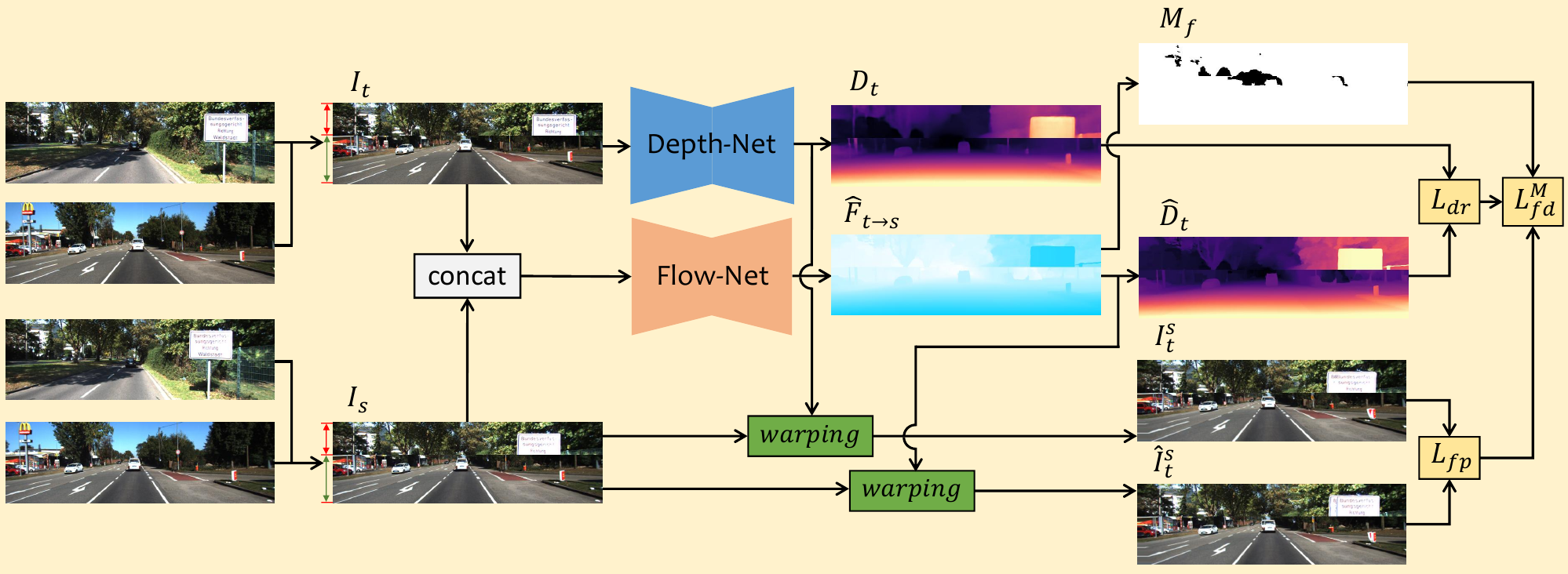}
    \caption{\textbf{Framework illustration.} Given a real stereo pair ($I_{t}$, $I_{s}$) that is refactored from training data by data grafting~\cite{EPCDepth}, the Flow-Net pretrained on synthetic data infers the prior flow $\hat{F}_{t\rightarrow s}$ that can then be converted to the pesudo depth lable $\hat{D}_{t}$ and the mask $M_{f}$ that removes those pixels out of estimation range. Multi-scale depth maps are estimated by the Depth-Net from $I_{t}$ and here we only draw the maximum scale depth map $D_{t}$ as an example. $I^{s}_{t}$ and $\hat{I}^{s}_{t}$ are synthesized from $I_{s}$ using $D_{t}$ and $\hat{D}_{t}$ respectively by inverse warping. Depth regression loss $L_{dr}$ is computed between $D_{t}$ and $\hat{D}_{t}$. Flow-guided photometric loss $L_{fp}$ is computed between $I^{s}_{t}$ and $\hat{I}^{s}_{t}$. Finally, total training loss $L^{M}_{fd}$ is calculated using $L_{dr}$, $L_{fp}$ and $M_{f}$.}\label{framework}
\end{figure*}

In this paper, we propose a flow distillation loss to replace the typical photometric loss (introduced in Sec.~\ref{lp}), a prior flow based mask to remove pixels out of the estimation range for self-supervised depth estimation networks. Our goal is to train depth networks using stereo pairs and constrain them with our proposed flow distillation loss and prior flow based mask. The pipeline of our framework is shown in Fig.~\ref{framework}. In the following subsections, we first introduce the typical photometric loss and automatic mask in Sec.~\ref{lp}, then describe our flow distillation loss in Sec~\ref{lfd}. and prior flow based mask in Sec~\ref{mask_section}. 

\subsection{Photometric Loss and Automatic Mask}\label{lp}

Previous stereo-based self-supervised works typically use photometric loss $L_{p}$ to train the model, assuming that the surface is Lambertian and has no occlusion~\cite{SfMLearner}. $L_{p}$ is between target frame $I_{t}$ and synthesized 
frame $I^{s}_{t}$ which is interpolated on source frame $I_{s}$ using predicted depth and relative camera pose, and is defined as: 
\begin{equation}
  \begin{aligned}
    L_{p}&=pe(I_{t}, I^{s}_{t}) \\
    &=\frac{\alpha}{2}(1-SSIM(I_{t},I^{s}_{t}))+(1-\alpha)\|I_{t}-I^{s}_{t}\|_{1}
  \end{aligned}
  \footnote{SSIM~\cite{SSIM} is computed over a $3\times 3$ pixel window, and $\alpha=0.85$.}
\end{equation}
Also widely used is the automatic mask $M_{p}$ for occlusion proposed in \cite{Monodepth2}, which is formulated as:
\begin{equation}
  M_{p}=\left[pe(I_{t}, I^{s}_{t})< pe(I_{t}, I_{s})\right]
  \footnote{[] here is the Iverson bracket.}
\end{equation}
It is still difficult to minimize the $L_{p}$ in real data. This is because there are multiple local minima with similar magnitudes, especially in regions of low texture and not fulfilling the assumption of color consistency~\cite{Depth_VO_Feat}. And the $L_{p}$ is disturbed by occluded pixels. But it's hard to remove occluded pixels completely by $M_{p}$ based on a simple comparison of geometry relationships. Also, removing occluded pixels means less supervision information. Furthermore, pixels out of estimation range can inhibit training but are usually neglected by previous methods. To avoid the above problems, we propose a flow distillation loss and a prior flow based mask. Our framework can achieve better performance due to an easier optimized loss function and more reliable supervision information.

\begin{figure}[t]
    \centering
    \subfigure[]{
      \includegraphics[width=8cm]{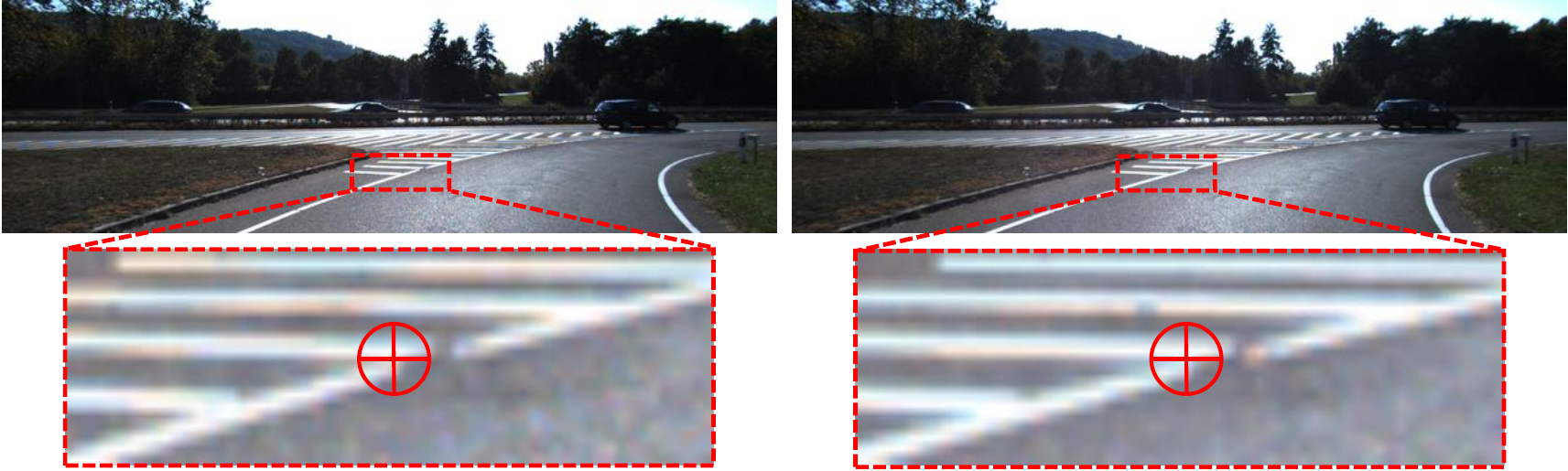}
      }
    \subfigure[]{
      \includegraphics[width=8cm]{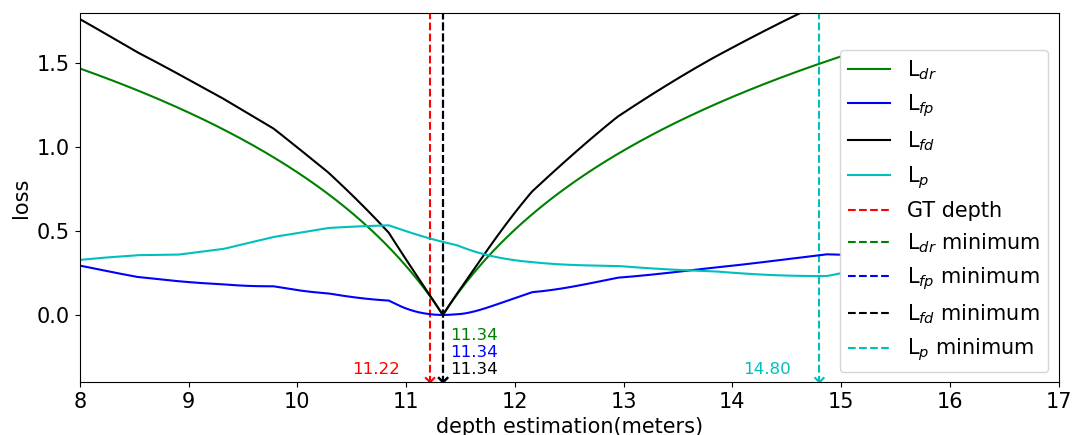}
      }
    \caption{
      \textbf{Loss visualization.} 
      (a)A pair of matching pixels on the left image and the right image. 
      (b)The relationship between the loss and the depth of matching points in the left subplot (a).
      The flow distillation loss is easier for optimization and has a more accurate global minimum when compared with photometric loss.
      }\label{loss_comparison}
\end{figure}

\subsection{Flow Distillation Loss}\label{lfd}
The flow distillation loss $L_{fd}$ consists of depth regression loss $L_{dr}$ and flow-guided photometric loss $L_{fp}$:
\begin{equation}
    L_{fd} = L_{dr}+L_{fp}
\end{equation}

The $L_{dr}$ is given by 
\begin{equation}
    L_{dr} = log(|D_{t}-\hat{D}_{t}|+1)
\end{equation}

The $D_{t}$ is the depth estimation and the pesudo depth lable $\hat{D}_{t}$ is computed from 
\begin{equation}
    \hat{D}_{t}=\frac{f_{x}b}{|\hat{F}_{t\rightarrow s}|}
\end{equation}
where $f_{x}$ is the focal length of the camera and $b$ is the baseline of the stereo. 

And the $L_{fp}$ is adopted to express the discrepancy between the reconstructions from $I_{s}$ respectively using $D_{t}$ and $\hat{D}_{t}$:
\begin{equation}
    \left\{
    \begin{array}{l}
        L_{fp} = |I^{s}_{t}-\hat{I}^{s}_{t}| \\
        I^{s}_{t}=f_{w}(I_{s}, D_{t}) \\
        \hat{I}^{s}_{t}=f_{w}(I_{s}, \hat{D}_{t})
    \end{array}
    \right.
\end{equation} 
where $f_{w}$ is the differentiable inverse warping operation.

The inspiration behind flow-guided photometric loss $L_{fp}$ is that the warping procedure in computing photometric loss is based on the flow which can be synthesized from depth estimation or directly predicted by pretrained optical flow estimation network. Optimization of $L_{fp}$ can be easier to reach better global minima because when $L_{fp}$ reaches minima, synthesized flow is closed to the predicted flow. And pretrained optical flow estimation network can predict enough accurate flow, so depth estimation can be more closed to ground truth after optimization of $L_{fp}$. By contrast, typical $L_{p}$ is harder to be optimized because of illumination changes.

The intuitive display is shown in Fig.~\ref{loss_comparison}. The red pixels marked in Fig.~\ref{loss_comparison}(a) are the pixel pairs matched by the stereo pair. Fig.~\ref{loss_comparison}(b) shows the loss curve for optimizing the depth of this matching point with depth regression loss $L_{dr}$, flow-guided photometric loss $L_{fp}$, flow distillation loss $L_{fd}$ and photometric loss $L_{p}$, respectively. There is the same global minimum for $L_{dr}$, $L_{fp}$, and $L_{fd}$, which is almost identical to the ground truth. And the $L_{fd}$ curve is steeper than $L_{dr}$ and $L_{fp}$, so it makes optimization of $L_{fd}$ easier. For $L_{p}$, there are multiple minimum points, where the optimal point fails to learn the correct depth value. Therefore, the $L_{fd}$ is easier for optimization and has a more accurate global minimum, when compared with the $L_{p}$.

\begin{figure}[h]
    \centering
    \includegraphics[scale=0.35]{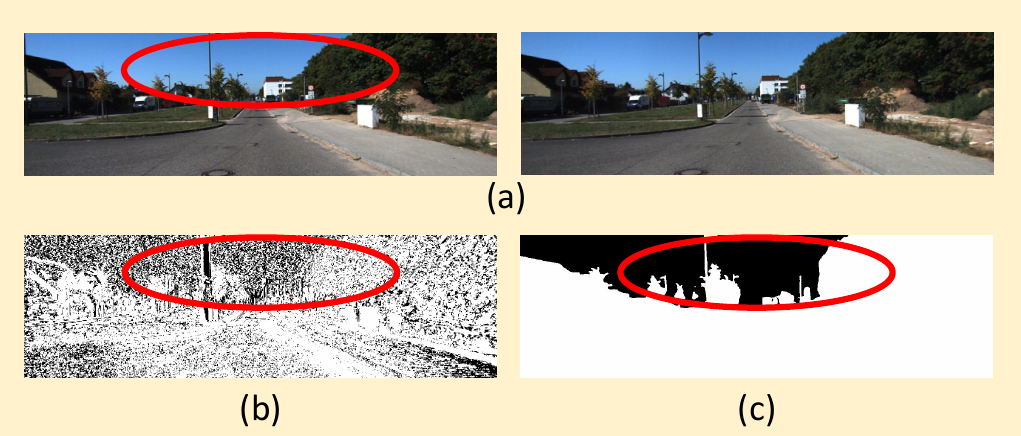}
    \caption{\textbf{Masks visualization.} 
      (a)Input target frame and source frame. 
      (b)Auto-mask $M_{p}$ proposed in \cite{Monodepth2}. 
      (c)Prior flow based mask.}\label{mask_visualization}
\end{figure}

\subsection{Prior Flow based Mask}\label{mask_section}

We use a prior flow based mask $M_{f}$ to remove those pixels out of range by checking the length of prior rigid flow. The mask value $M_{f}(p_{i})$ of the pixel at position $p_i$ can be formulated as:
\begin{equation}\label{mask_eqn}
    \left\{
      \begin{array}{l}
      \begin{split}
      &M_{f}(p_{i}) = \left\{
      \begin{array}{ll}
        1      & ,\ p_{i}\in V \\
        0      & ,\ else \\
      \end{array}
      \right.  \\
      &V = \left\{p_{t}\bigg| |\hat{F}_{t\rightarrow s}(p_{i}) | > \frac{f_{x}b}{\delta}\right\}
      \end{split}
      \end{array}
    \right.
\end{equation}
where $\hat{F}_{t\rightarrow s}$ denotes the prior flow from target frame to source frame and $\delta$ are set to 80.

In depth estimation, out-of-range depths (greater than 80m in KITTI) drop dramatically in accuracy. In previous works, masks do not remove all those pixels out-of-range. So, noise is brought in photometric loss because those out-of-range pixels always fail to match corresponding pixels in the warping procedure. Visualization results in Fig.~\ref{mask_visualization} intuitively show that compared with automatic mask $M_{p}$ proposed in~\cite{Monodepth2}, our mask $M_{f}$ can filter out out-of-range pixels more completely, making it more stable and less susceptible to noise interference.

\subsection{Final Training Loss}
We combine the flow distillation loss and prior flow based mask as:
\begin{equation}
  L^{M}_{fd}=\frac{1}{T}\sum_{i}M_{f}(p_{i})L_{fd}(p_{i})
\end{equation}
where $T$ denotes the number of pixels reserved by the mask, and average over each scale.

\subsection{Network Architecture}
We implement the Flow-Net with RAFT-Stereo~\cite{RAFT_Stereo} which is based on GRU~\cite{cho2014properties} and has excellent accuracy and good generalization ability. For simplicity, we directly use the official model\footnote{\url{https://github.com/princeton-vl/RAFT-Stereo}} that is pretrained on Scene Flow dataset~\cite{SceneFlow}.

For the Depth-Net, we use the same architecture as~\cite{EPCDepth} which uses ResNet18 as backbone and RSU block as the bridge between different scale features and disparity prediction blocks to output full-scale predictions. The outputs $\sigma$ of the prediction blocks are further constrained between 0.1 and 80 units with $D=1/(a\sigma+b)$.

\section{Experiments}

In this section, we evaluate our proposed model on the KITTI dataset~\cite{KITTI} to verify its state-of-art 
performance and we validate the generalization ability of our model on the NYU-Depth-v2 dataset~\cite{NYUV2}. 
Furthermore, we conduct an ablation study to demonstrate the effectiveness of our contributions.

\subsection{Datasets}
\paragraph{KITTI} 
KITTI dataset was captured by a driving vehicle with cameras and depth sensors around the mid-size city of Karlsruhe, in rural areas, and on highways. It is widely used for outdoor monocular depth estimation and we use the Eigen split~\cite{Eigen} that consists of 22600 stereo image pairs for training and 697 images for testing. The train set is from 32 scenes and the test set is from other 29 scenes. 
\paragraph{NYU-Depth-v2} 
NYU-Depth-v2 dataset was collected with a Kinect sensor in total of 582 indoor scenes. To validate the generalization ability of our model, we use the official test set that consists of 654 images with depth GTs.

\subsection{Inplementation Details}\label{inplementation_details}
Our work is implemented in PyTorch on one Nvidia Tesla V100 GPU. For training, we use the Adam optimizer~\cite{Adam}($\beta_{1}$ = 0.9, $\beta_{2}$ = 0.999). The total number of epochs is set to 20 with a batch size of 12 and an input/output resolution of $192\times 640$ unless otherwise specified. The initial learning rate is $1\times10^{-4}$ and decays after the 10th epoch with a factor of $0.1$. For evaluation, we resize the estimated depth map to the ground-truth depth resolution using bilinear interpolation.

\begin{table*}[t]
    \caption{\textbf{Quantitative results on the KITTI dataset using Eigen split~\cite{Eigen}.} For \textcolor{red!30}{\textbf{red}} 
    metrics, lower is better. And higher is better for \textcolor{blue!30}{\textbf{blue}} metrics. PP represents post-processing~\cite{Monodepth}. In data column, D refers to the methods supervised by the ground truth depth, M means that the supervision is from monocular video, S means that the supervision is from stereo pairs, C$^{\dag}$ means using predicted segmentation lables and S$^{*}$ means using extra synthetic sceneflow dataset. The best results are in \textbf{bold}.
    }\label{kitti_quantitative_result}
    \centering
    \footnotesize
    \setlength\tabcolsep{3pt}
    \begin{tabular}{lcccccccccc}
    \hline
    \multicolumn{1}{l}{Method} & \multicolumn{1}{|c}{PP} & \multicolumn{1}{|c|}{Data} & \multicolumn{1}{c|}{Resolution} & \multicolumn{1}{c}{\cellcolor{red!30}Abs Rel} & \multicolumn{1}{c}{\cellcolor{red!30}Sq Rel} & \multicolumn{1}{c}{\cellcolor{red!30}RMSE} & \multicolumn{1}{c|}{\cellcolor{red!30}RMSE$_{log}$} & \multicolumn{1}{c}{\cellcolor{blue!30}$\delta^{1}$} & \multicolumn{1}{c}{\cellcolor{blue!30}$\delta^{2}$} & \multicolumn{1}{c}{\cellcolor{blue!30}$\delta^{3}$} \\ 
    \hline
    \multicolumn{1}{l}{DORN~\cite{DORN} (ResNet101)}  & \multicolumn{1}{|l}{}  & \multicolumn{1}{|l|}{D}           & \multicolumn{1}{c|}{$385\times 513$ crop} & \multicolumn{1}{c}{0.099}   & \multicolumn{1}{c}{0.593}          & \multicolumn{1}{c}{3.714}          & \multicolumn{1}{c|}{0.161}          & \multicolumn{1}{c}{0.897}          & \multicolumn{1}{c}{0.966} & \multicolumn{1}{c}{\textbf{0.986}} \\
    \multicolumn{1}{l}{BTS~\cite{BTS} (DenseNet-161)}  & \multicolumn{1}{|l}{}  & \multicolumn{1}{|l|}{D}           & \multicolumn{1}{c|}{$352\times 704$ crop} & \multicolumn{1}{c}{0.091}   & \multicolumn{1}{c}{0.555}          & \multicolumn{1}{c}{4.033}          & \multicolumn{1}{c|}{0.174}          & \multicolumn{1}{c}{0.904}          & \multicolumn{1}{c}{0.967} & \multicolumn{1}{c}{0.984} \\
    \multicolumn{1}{l}{AdaBins~\cite{AdaBins} (EfficientNet-B5)}  & \multicolumn{1}{|l}{}  & \multicolumn{1}{|l|}{D}           & \multicolumn{1}{c|}{$352\times 704$ crop} & \multicolumn{1}{c}{0.086}   & \multicolumn{1}{c}{0.475}          & \multicolumn{1}{c}{3.621}          & \multicolumn{1}{c|}{0.167}          & \multicolumn{1}{c}{0.918}          & \multicolumn{1}{c}{0.970} & \multicolumn{1}{c}{0.985} \\
    \multicolumn{1}{l}{NeWCRFs~\cite{NeWCRFs} (swin-ViT)}  & \multicolumn{1}{|l}{}  & \multicolumn{1}{|l|}{D}           & \multicolumn{1}{c|}{$352\times 1120$ crop} & \multicolumn{1}{c}{\textbf{0.080}} & \multicolumn{1}{c}{\textbf{0.426}} & \multicolumn{1}{c}{\textbf{3.400}} & \multicolumn{1}{c|}{\textbf{0.158}} & \multicolumn{1}{c}{\textbf{0.930}} & \multicolumn{1}{c}{\textbf{0.973}} & \multicolumn{1}{c}{\textbf{0.986}} \\
    \hline
    \hline
    \multicolumn{1}{l}{MonoDepth2~\cite{Monodepth2}}  & \multicolumn{1}{|l}{\textcolor{red}{\ding{51}}}  & \multicolumn{1}{|l|}{MS}           & \multicolumn{1}{c|}{$192\times 640$} & \multicolumn{1}{c}{0.104}   & \multicolumn{1}{c}{0.786}          & \multicolumn{1}{c}{4.687}          & \multicolumn{1}{c|}{0.194}          & \multicolumn{1}{c}{0.876}          & \multicolumn{1}{c}{0.958}          & \multicolumn{1}{c}{0.980}            \\ 
    \multicolumn{1}{l}{DepthHints~\cite{Depth_Hints}} & \multicolumn{1}{|l}{\textcolor{red}{\ding{51}}}   & \multicolumn{1}{|l|}{MS}           & \multicolumn{1}{c|}{$192\times 640$} & \multicolumn{1}{c}{0.105}   & \multicolumn{1}{c}{0.769}          & \multicolumn{1}{c}{4.627}          & \multicolumn{1}{c|}{0.189}          & \multicolumn{1}{c}{0.875}          & \multicolumn{1}{c}{0.959}          & \multicolumn{1}{c}{0.982}            \\ 
    \multicolumn{1}{l}{HR-Depth~\cite{HR_Depth}}  & \multicolumn{1}{|l}{}  & \multicolumn{1}{|l|}{MS}           & \multicolumn{1}{c|}{$192\times 640$} & \multicolumn{1}{c}{0.107}   & \multicolumn{1}{c}{0.785}          & \multicolumn{1}{c}{4.612}          & \multicolumn{1}{c|}{0.185}          & \multicolumn{1}{c}{0.887}          & \multicolumn{1}{c}{0.962}          & \multicolumn{1}{c}{0.982}            \\ 
    \multicolumn{1}{l}{CADepth-Net~\cite{CADepth-Net} (ResNet50)}  & \multicolumn{1}{|l}{}  & \multicolumn{1}{|l|}{MS}           & \multicolumn{1}{c|}{$192\times 640$} & \multicolumn{1}{c}{\textbf{0.102}}   & \multicolumn{1}{c}{\textbf{0.752}}          & \multicolumn{1}{c}{\textbf{4.502}}          & \multicolumn{1}{c|}{\textbf{0.181}}          & \multicolumn{1}{c}{\textbf{0.894}}          & \multicolumn{1}{c}{\textbf{0.964}} & \multicolumn{1}{c}{\textbf{0.983}} \\ 
    \hline
    \multicolumn{1}{l}{Guo et al.~\cite{Guo_2018_ECCV} w/o Fintuned (VGG-16)}  & \multicolumn{1}{|l}{\textcolor{red}{\ding{51}}}    & \multicolumn{1}{|l|}{SS$^{*}$}           & \multicolumn{1}{c|}{$384\times 1280$} & \multicolumn{1}{c}{0.109}          & \multicolumn{1}{c}{0.822}          & \multicolumn{1}{c}{4.656}          & \multicolumn{1}{c|}{0.192}          & \multicolumn{1}{c}{0.868}          & \multicolumn{1}{c}{0.958}          & \multicolumn{1}{c}{0.981}            \\ 
    \multicolumn{1}{l}{Guo et al.~\cite{Guo_2018_ECCV} Fintuned (VGG-16)}  & \multicolumn{1}{|l}{\textcolor{red}{\ding{51}}}    & \multicolumn{1}{|l|}{SS$^{*}$}           & \multicolumn{1}{c|}{$384\times 1280$} & \multicolumn{1}{c}{0.099}          & \multicolumn{1}{c}{0.745}          & \multicolumn{1}{c}{4.424}          & \multicolumn{1}{c|}{0.182}          & \multicolumn{1}{c}{0.884}          & \multicolumn{1}{c}{0.963}          & \multicolumn{1}{c}{0.983}            \\ 
    \multicolumn{1}{l}{MonoDepth2~\cite{Monodepth2}}  & \multicolumn{1}{|l}{\textcolor{red}{\ding{51}}}  & \multicolumn{1}{|l|}{S}           & \multicolumn{1}{c|}{$192\times 640$} & \multicolumn{1}{c}{0.108}   & \multicolumn{1}{c}{0.842}          & \multicolumn{1}{c}{4.891}          & \multicolumn{1}{c|}{0.207}          & \multicolumn{1}{c}{0.866}          & \multicolumn{1}{c}{0.949}          & \multicolumn{1}{c}{0.976}            \\ 
    \multicolumn{1}{l}{DepthHints~\cite{Depth_Hints}} & \multicolumn{1}{|l}{\textcolor{red}{\ding{51}}}  & \multicolumn{1}{|l|}{S}   & \multicolumn{1}{c|}{$192\times 640$}  & \multicolumn{1}{c}{0.106}          & \multicolumn{1}{c}{0.780}          & \multicolumn{1}{c}{4.695}          & \multicolumn{1}{c|}{0.193}          & \multicolumn{1}{c}{0.875}          & \multicolumn{1}{c}{0.958}   & \multicolumn{1}{c}{0.980}   \\ 
    \multicolumn{1}{l}{EPCDepth~\cite{EPCDepth}}  & \multicolumn{1}{|l}{\textcolor{red}{\ding{51}}}    & \multicolumn{1}{|l|}{S}           & \multicolumn{1}{c|}{$192\times 640$}  & \multicolumn{1}{c}{0.099}          & \multicolumn{1}{c}{0.754}          & \multicolumn{1}{c}{4.490}          & \multicolumn{1}{c|}{0.183}          & \multicolumn{1}{c}{0.888}          & \multicolumn{1}{c}{0.963}   & \multicolumn{1}{c}{0.982}   \\ 
    \multicolumn{1}{l}{\textbf{Ours}} & \multicolumn{1}{|l}{\textcolor{red}{\ding{51}}} & \multicolumn{1}{|l|}{SS$^{*}$} & \multicolumn{1}{c|}{$192\times 640$} & \multicolumn{1}{c}{\textbf{0.093}} & \multicolumn{1}{c}{\textbf{0.634}} & \multicolumn{1}{c}{\textbf{4.123}} & \multicolumn{1}{c|}{\textbf{0.174}} & \multicolumn{1}{c}{\textbf{0.900}} & \multicolumn{1}{c}{\textbf{0.967}} & \multicolumn{1}{c}{\textbf{0.984}} \\ 
    \hline 
    \hline
    \multicolumn{1}{l}{MonoDepth2~\cite{Monodepth2}}  & \multicolumn{1}{|l}{\textcolor{red}{\ding{51}}}  & \multicolumn{1}{|l|}{MS}           & \multicolumn{1}{c|}{$320\times 1024$} & \multicolumn{1}{c}{0.104}   & \multicolumn{1}{c}{0.775}          & \multicolumn{1}{c}{4.562}          & \multicolumn{1}{c|}{0.191}          & \multicolumn{1}{c}{0.878}          & \multicolumn{1}{c}{0.959}          & \multicolumn{1}{c}{0.981}            \\ 
    \multicolumn{1}{l}{DepthHints~\cite{Depth_Hints}} & \multicolumn{1}{|l}{\textcolor{red}{\ding{51}}}   & \multicolumn{1}{|l|}{MS}           & \multicolumn{1}{c|}{$320\times 1024$} & \multicolumn{1}{c}{0.098}   & \multicolumn{1}{c}{0.702}          & \multicolumn{1}{c}{4.398}          & \multicolumn{1}{c|}{0.183}          & \multicolumn{1}{c}{0.887}          & \multicolumn{1}{c}{0.963}          & \multicolumn{1}{c}{0.983}            \\ 
    \multicolumn{1}{l}{HR-Depth~\cite{HR_Depth}}  & \multicolumn{1}{|l}{}  & \multicolumn{1}{|l|}{MS}           & \multicolumn{1}{c|}{$320\times 1024$} & \multicolumn{1}{c}{0.101}   & \multicolumn{1}{c}{0.716}          & \multicolumn{1}{c}{4.395}          & \multicolumn{1}{c|}{0.179}          & \multicolumn{1}{c}{0.899}          & \multicolumn{1}{c}{0.966}          & \multicolumn{1}{c}{0.983}            \\ 
    \multicolumn{1}{l}{Feat-Depth~\cite{Feat_Depth} (ResNet50)}  & \multicolumn{1}{|l}{}  & \multicolumn{1}{|l|}{MS}           & \multicolumn{1}{c|}{$320\times 1024$} & \multicolumn{1}{c}{0.099}   & \multicolumn{1}{c}{0.697}          & \multicolumn{1}{c}{4.427}          & \multicolumn{1}{c|}{0.184}          & \multicolumn{1}{c}{0.889}          & \multicolumn{1}{c}{0.963}          & \multicolumn{1}{c}{0.982}            \\ 
    \multicolumn{1}{l}{CADepth-Net~\cite{CADepth-Net} (ResNet50)}  & \multicolumn{1}{|l}{}  & \multicolumn{1}{|l|}{MS}           & \multicolumn{1}{c|}{$320\times 1024$} & \multicolumn{1}{c}{\textbf{0.096}}   & \multicolumn{1}{c}{\textbf{0.694}} & \multicolumn{1}{c}{\textbf{4.264}} & \multicolumn{1}{c|}{\textbf{0.173}} & \multicolumn{1}{c}{\textbf{0.908}} & \multicolumn{1}{c}{\textbf{0.968}} & \multicolumn{1}{c}{\textbf{0.984}} \\ 
    \hline
    \multicolumn{1}{l}{MonoDepth2~\cite{Monodepth2}}  & \multicolumn{1}{|l}{\textcolor{red}{\ding{51}}}  & \multicolumn{1}{|l|}{S}           & \multicolumn{1}{c|}{$320\times 1024$} & \multicolumn{1}{c}{0.105}          & \multicolumn{1}{c}{0.822}          & \multicolumn{1}{c}{4.692}          & \multicolumn{1}{c|}{0.199}          & \multicolumn{1}{c}{0.874}          & \multicolumn{1}{c}{0.954}          & \multicolumn{1}{c}{0.977}            \\ 
    \multicolumn{1}{l}{DepthHints~\cite{Depth_Hints}}  & \multicolumn{1}{|l}{\textcolor{red}{\ding{51}}}   & \multicolumn{1}{|l|}{S} & \multicolumn{1}{c|}{$320\times 1024$} & \multicolumn{1}{c}{0.099}          & \multicolumn{1}{c}{0.723}          & \multicolumn{1}{c}{4.445}          & \multicolumn{1}{c|}{0.187}          & \multicolumn{1}{c}{0.886}          & \multicolumn{1}{c}{0.962}          & \multicolumn{1}{c}{0.981}            \\ 
    \multicolumn{1}{l}{EdgeDepth~\cite{Edge_Depth}}  & \multicolumn{1}{|l}{\textcolor{red}{\ding{51}}}   & \multicolumn{1}{|l|}{SC$^{\dag}$} & \multicolumn{1}{c|}{$320\times 1024$} & \multicolumn{1}{c}{0.097}          & \multicolumn{1}{c}{0.675}          & \multicolumn{1}{c}{4.350}          & \multicolumn{1}{c|}{0.180}          & \multicolumn{1}{c}{0.890}          & \multicolumn{1}{c}{0.965}          & \multicolumn{1}{c}{0.983}            \\
    \multicolumn{1}{l}{EPCDepth~\cite{EPCDepth}}  & \multicolumn{1}{|l}{\textcolor{red}{\ding{51}}}    & \multicolumn{1}{|l|}{S}           & \multicolumn{1}{c|}{$320\times 1024$} & \multicolumn{1}{c}{0.093}          & \multicolumn{1}{c}{0.671}          & \multicolumn{1}{c}{4.297}          & \multicolumn{1}{c|}{0.178}          & \multicolumn{1}{c}{0.899}          & \multicolumn{1}{c}{0.965}          & \multicolumn{1}{c}{0.983}            \\ 
    \multicolumn{1}{l}{\textbf{Ours}} & \multicolumn{1}{|l}{\textcolor{red}{\ding{51}}} & \multicolumn{1}{|l|}{SS$^{*}$}    & \multicolumn{1}{c|}{$320\times 1024$} & \multicolumn{1}{c}{\textbf{0.088}} & \multicolumn{1}{c}{\textbf{0.583}} & \multicolumn{1}{c}{\textbf{3.924}} & \multicolumn{1}{c|}{\textbf{0.168}} & \multicolumn{1}{c}{\textbf{0.909}} & \multicolumn{1}{c}{\textbf{0.970}} & \multicolumn{1}{c}{\textbf{0.985}}   \\
    \hline
    \multicolumn{1}{l}{DepthHints~\cite{Depth_Hints} (ResNet50)}  & \multicolumn{1}{|l}{\textcolor{red}{\ding{51}}}   & \multicolumn{1}{|l|}{S} & \multicolumn{1}{c|}{$320\times 1024$} & \multicolumn{1}{c}{0.096}          & \multicolumn{1}{c}{0.710}          & \multicolumn{1}{c}{4.393}          & \multicolumn{1}{c|}{0.185}          & \multicolumn{1}{c}{0.890}          & \multicolumn{1}{c}{0.962}          & \multicolumn{1}{c}{0.983}            \\ 
    \multicolumn{1}{l}{EdgeDepth~\cite{Edge_Depth} (ResNet50)}  & \multicolumn{1}{|l}{\textcolor{red}{\ding{51}}}   & \multicolumn{1}{|l|}{SC$^{\dag}$} & \multicolumn{1}{c|}{$320\times 1024$} & \multicolumn{1}{c}{0.091}          & \multicolumn{1}{c}{0.646}          & \multicolumn{1}{c}{4.244}          & \multicolumn{1}{c|}{0.177}          & \multicolumn{1}{c}{0.898}          & \multicolumn{1}{c}{0.966}          & \multicolumn{1}{c}{0.983}            \\
    \multicolumn{1}{l}{EPCDepth~\cite{EPCDepth} (ResNet50)}  & \multicolumn{1}{|l}{\textcolor{red}{\ding{51}}}    & \multicolumn{1}{|l|}{S}           & \multicolumn{1}{c|}{$320\times 1024$} & \multicolumn{1}{c}{0.091}          & \multicolumn{1}{c}{0.646}          & \multicolumn{1}{c}{4.207}          & \multicolumn{1}{c|}{0.176}          & \multicolumn{1}{c}{0.901}          & \multicolumn{1}{c}{0.966}          & \multicolumn{1}{c}{0.983}            \\ 
    \multicolumn{1}{l}{\textbf{Ours (ResNet50)}} & \multicolumn{1}{|l}{\textcolor{red}{\ding{51}}} & \multicolumn{1}{|l|}{SS$^{*}$}    & \multicolumn{1}{c|}{$320\times 1024$} & \multicolumn{1}{c}{\textbf{0.086}} & \multicolumn{1}{c}{\textbf{0.575}} & \multicolumn{1}{c}{\textbf{3.873}} & \multicolumn{1}{c|}{\textbf{0.166}} & \multicolumn{1}{c}{\textbf{0.910}} & \multicolumn{1}{c}{\textbf{0.971}} & \multicolumn{1}{c}{\textbf{0.985}}   \\
    \hline
    \end{tabular}
\end{table*}

\begin{figure*}[h]
    \centering
    \subfigure{
        \begin{minipage}[t]{0.17\textwidth}
            \centering
            {\scriptsize{Input}}
        \end{minipage}
        \begin{minipage}[t]{0.17\textwidth}
            \centering
            {\scriptsize{Monodepth2~\cite{Monodepth2}}}
        \end{minipage}
        \begin{minipage}[t]{0.17\textwidth}
            \centering
            {\scriptsize{DepthHints~\cite{Depth_Hints}}}
        \end{minipage}
        \begin{minipage}[t]{0.17\textwidth}
            \centering
            {\scriptsize{EPCDepth~\cite{EPCDepth}}}
        \end{minipage}
        \begin{minipage}[t]{0.17\textwidth}
            \centering
            {\scriptsize{Ours(FG-Depth)}}
        \end{minipage}
    }\vspace{-3mm}
    \subfigure{
        \begin{minipage}[t]{0.17\linewidth}
            \centering
            \raisebox{-0.15cm}{\includegraphics[width=3cm]{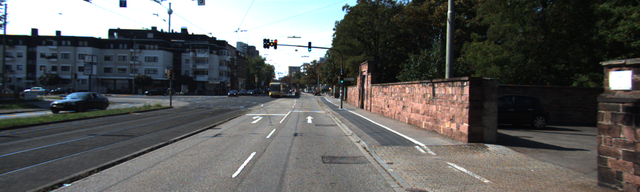}}
        \end{minipage}
        \begin{minipage}[t]{0.17\linewidth}
            \centering
            \raisebox{-0.15cm}{\includegraphics[width=3cm]{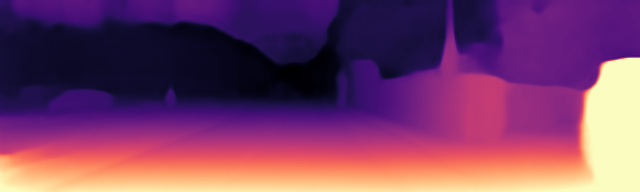}}
        \end{minipage}
        \begin{minipage}[t]{0.17\linewidth}
            \centering
            \raisebox{-0.15cm}{\includegraphics[width=3cm]{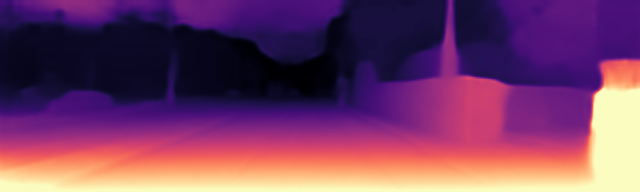}}
        \end{minipage}
        \begin{minipage}[t]{0.17\linewidth}
            \centering
            \raisebox{-0.15cm}{\includegraphics[width=3cm]{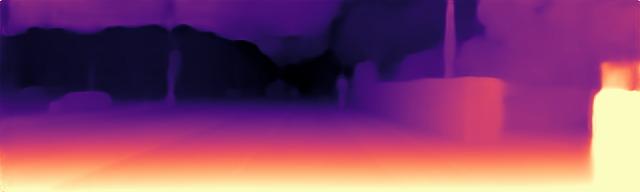}}
        \end{minipage}
        \begin{minipage}[t]{0.17\linewidth}
            \centering
            \raisebox{-0.15cm}{\includegraphics[width=3cm]{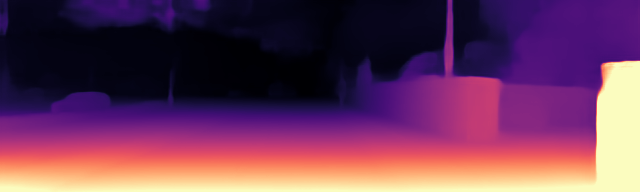}}
        \end{minipage}
    }\vspace{-2mm}
    \subfigure{
        \begin{minipage}[t]{0.17\linewidth}
            \centering
            \raisebox{-0.15cm}{\includegraphics[width=3cm]{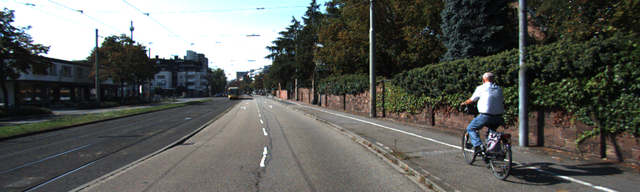}}
        \end{minipage}
        \begin{minipage}[t]{0.17\linewidth}
            \centering
            \raisebox{-0.15cm}{\includegraphics[width=3cm]{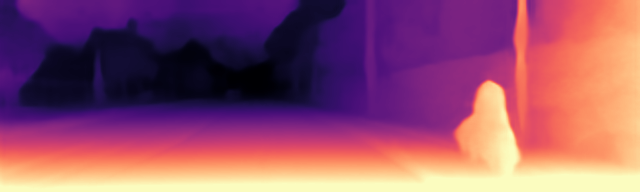}}
        \end{minipage}
        \begin{minipage}[t]{0.17\linewidth}
            \centering
            \raisebox{-0.15cm}{\includegraphics[width=3cm]{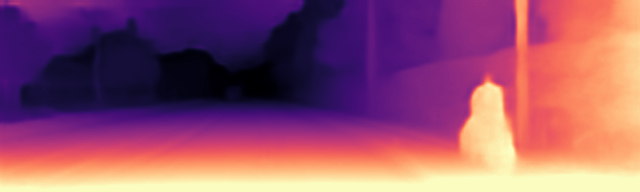}}
        \end{minipage}
        \begin{minipage}[t]{0.17\linewidth}
            \centering
            \raisebox{-0.15cm}{\includegraphics[width=3cm]{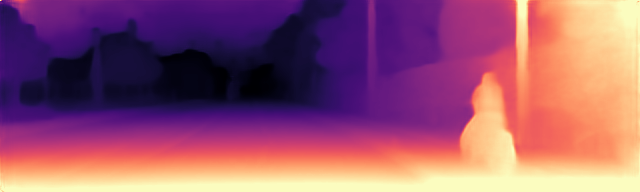}}
        \end{minipage}
        \begin{minipage}[t]{0.17\linewidth}
            \centering
            \raisebox{-0.15cm}{\includegraphics[width=3cm]{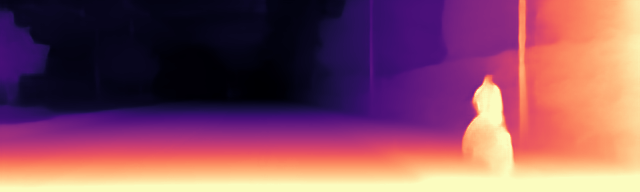}}
        \end{minipage}
    }\vspace{-2mm}
    \subfigure{
        \begin{minipage}[t]{0.17\linewidth}
            \centering
            \raisebox{-0.15cm}{\includegraphics[width=3cm]{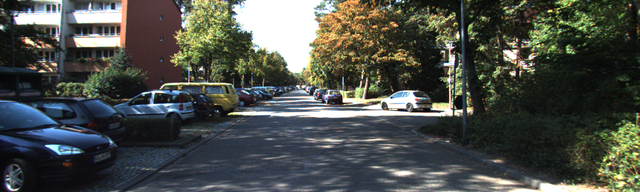}}
        \end{minipage}
        \begin{minipage}[t]{0.17\linewidth}
            \centering
            \raisebox{-0.15cm}{\includegraphics[width=3cm]{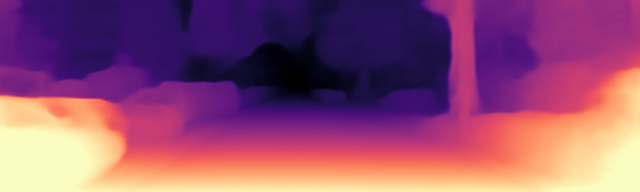}}
        \end{minipage}
        \begin{minipage}[t]{0.17\linewidth}
            \centering
            \raisebox{-0.15cm}{\includegraphics[width=3cm]{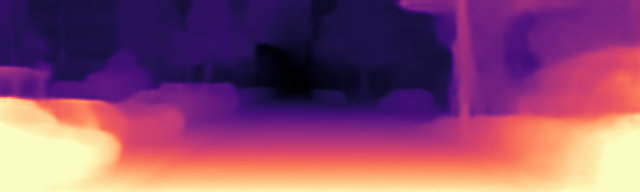}}
        \end{minipage}
        \begin{minipage}[t]{0.17\linewidth}
            \centering
            \raisebox{-0.15cm}{\includegraphics[width=3cm]{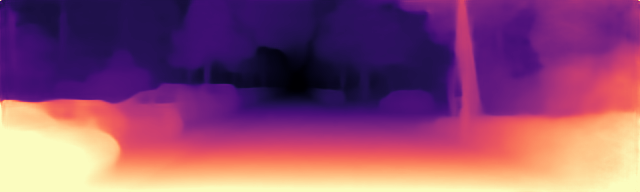}}
        \end{minipage}
        \begin{minipage}[t]{0.17\linewidth}
            \centering
            \raisebox{-0.15cm}{\includegraphics[width=3cm]{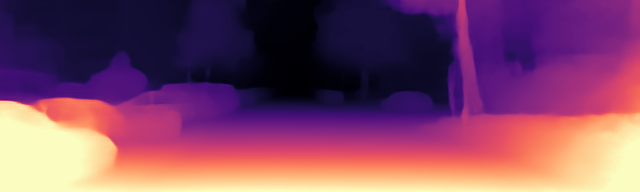}}
        \end{minipage}
    }\vspace{-2mm}
    \subfigure{
        \begin{minipage}[t]{0.17\linewidth}
            \centering
            \raisebox{-0.15cm}{\includegraphics[width=3cm]{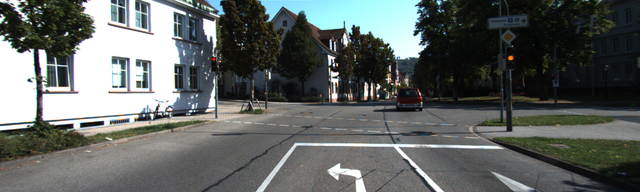}}
        \end{minipage}
        \begin{minipage}[t]{0.17\linewidth}
            \centering
            \raisebox{-0.15cm}{\includegraphics[width=3cm]{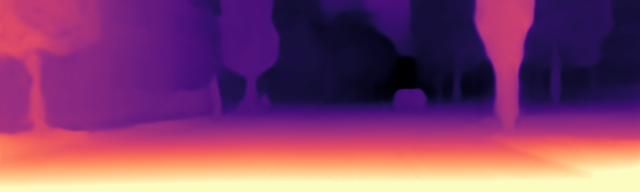}}
        \end{minipage}
        \begin{minipage}[t]{0.17\linewidth}
            \centering
            \raisebox{-0.15cm}{\includegraphics[width=3cm]{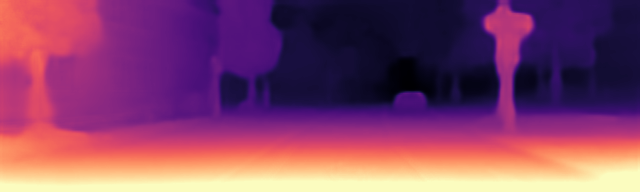}}
        \end{minipage}
        \begin{minipage}[t]{0.17\linewidth}
            \centering
            \raisebox{-0.15cm}{\includegraphics[width=3cm]{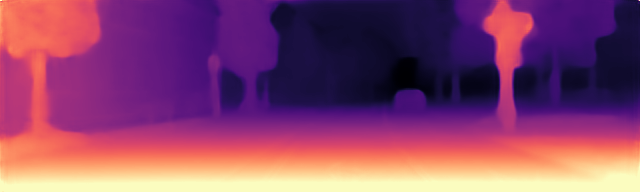}}
        \end{minipage}
        \begin{minipage}[t]{0.17\linewidth}
            \centering
            \raisebox{-0.15cm}{\includegraphics[width=3cm]{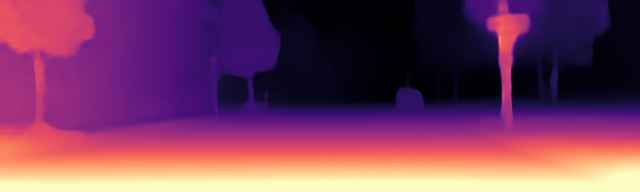}}
        \end{minipage}
    }\vspace{-2mm}
    \subfigure{
        \begin{minipage}[t]{0.17\linewidth}
            \centering
            \raisebox{-0.15cm}{\includegraphics[width=3cm]{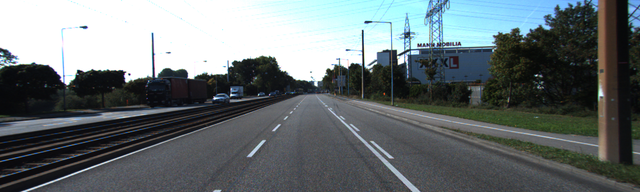}}
        \end{minipage}
        \begin{minipage}[t]{0.17\linewidth}
           \centering
            \raisebox{-0.15cm}{\includegraphics[width=3cm]{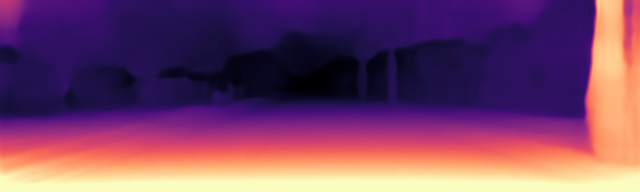}}
        \end{minipage}
        \begin{minipage}[t]{0.17\linewidth}
            \centering
            \raisebox{-0.15cm}{\includegraphics[width=3cm]{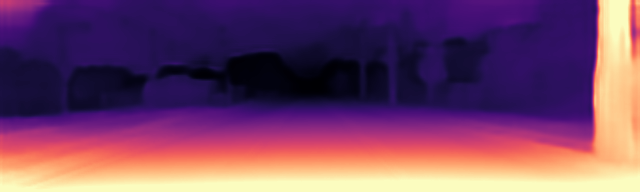}}
        \end{minipage}
        \begin{minipage}[t]{0.17\linewidth}
            \centering
            \raisebox{-0.15cm}{\includegraphics[width=3cm]{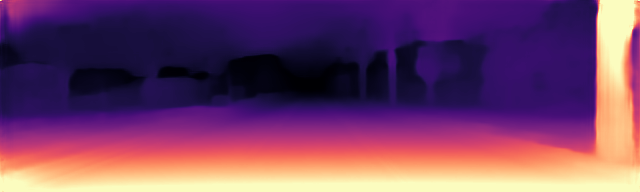}}
        \end{minipage}
        \begin{minipage}[t]{0.17\linewidth}
            \centering
            \raisebox{-0.15cm}{\includegraphics[width=3cm]{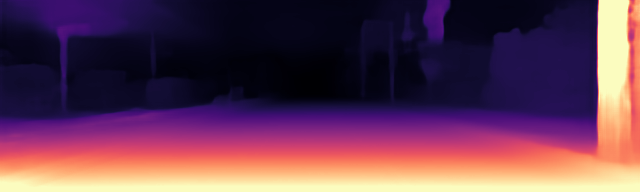}}
        \end{minipage}
    }\vspace{-2mm}
    \subfigure{
        \begin{minipage}[t]{0.17\linewidth}
            \centering
            \raisebox{-0.15cm}{\includegraphics[width=3cm]{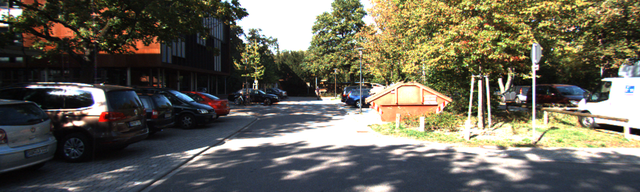}}
        \end{minipage}
        \begin{minipage}[t]{0.17\linewidth}
            \centering
            \raisebox{-0.15cm}{\includegraphics[width=3cm]{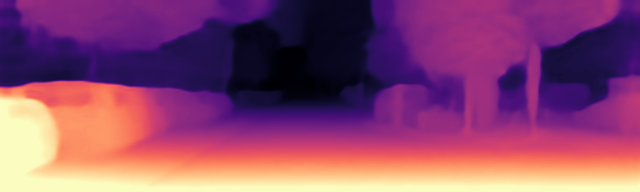}}
        \end{minipage}
        \begin{minipage}[t]{0.17\linewidth}
            \centering
            \raisebox{-0.15cm}{\includegraphics[width=3cm]{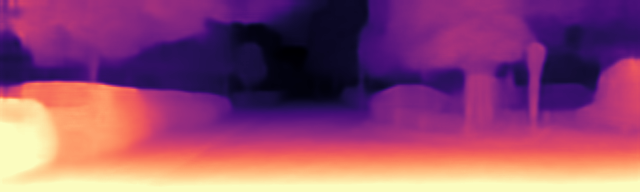}}
        \end{minipage}
        \begin{minipage}[t]{0.17\linewidth}
            \centering
            \raisebox{-0.15cm}{\includegraphics[width=3cm]{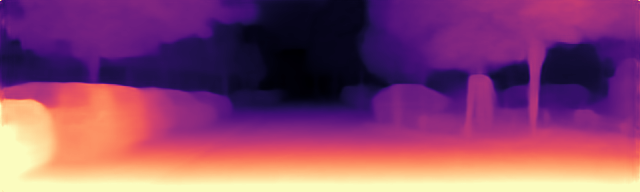}}
        \end{minipage}
        \begin{minipage}[t]{0.17\linewidth}
            \centering
            \raisebox{-0.15cm}{\includegraphics[width=3cm]{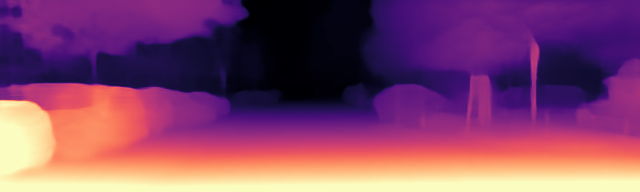}}
        \end{minipage}
    }\vspace{-2mm}
    \subfigure{
        \begin{minipage}[t]{0.17\linewidth}
            \centering
            \raisebox{-0.15cm}{\includegraphics[width=3cm]{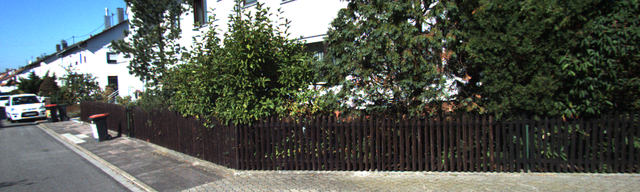}}
        \end{minipage}
        \begin{minipage}[t]{0.17\linewidth}
            \centering
            \raisebox{-0.15cm}{\includegraphics[width=3cm]{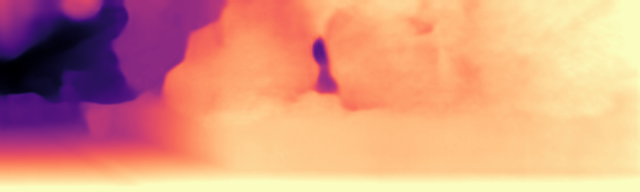}}
        \end{minipage}
        \begin{minipage}[t]{0.17\linewidth}
            \centering
            \raisebox{-0.15cm}{\includegraphics[width=3cm]{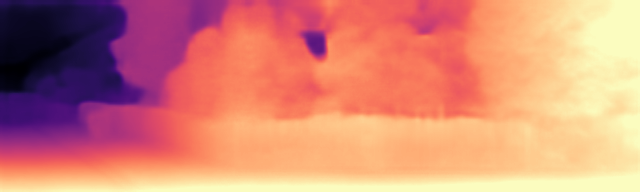}}
        \end{minipage}
        \begin{minipage}[t]{0.17\linewidth}
            \centering
            \raisebox{-0.15cm}{\includegraphics[width=3cm]{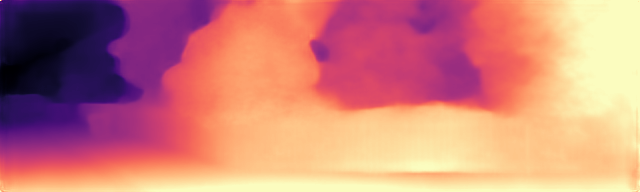}}
        \end{minipage}
        \begin{minipage}[t]{0.17\linewidth}
            \centering
            \raisebox{-0.15cm}{\includegraphics[width=3cm]{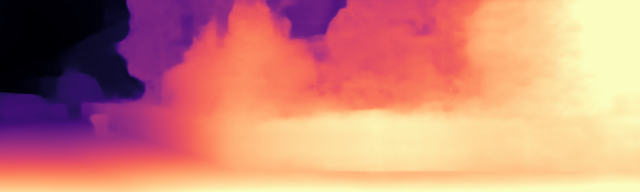}}
        \end{minipage}
    }\vspace{-2mm}
    \caption{\textbf{Qualitative results on the KITTI dataset using Eigen split.} Our model, FG-Depth, produces the sharpest results even in low-texture regions and on thin structures.}\label{kitti_qualitative_result}
\end{figure*}

With a 50\% chance, we flip the input images horizontally, apply data grafting~\cite{EPCDepth}  with the same setting as~\cite{EPCDepth} and add color augmentations where we perform random brightness, contrast, saturation, and hue jitter by sampling uniform distributions in ranges of [0.8,1.2], [0.8,1.2], [0.8,1.2], [0.9,1.1] respectively. The color augmentations are applied to the images that are fed to the Depth-Net rather than those fed to the Flow-Net and the loss function. 

\subsection{Depth Estimation Performance}
Firstly, we verify the performance of our model on the KITTI dataset. For a fair comparison, we use the metrics proposed in \cite{Eigen} with Garg's crop~\cite{Garg} and a standard distance cap of 80 meters. The same as other comparative self-supervised methods, we use the same post-processing steps as theirs~\cite{Monodepth}. The quantitative results are summarized in Tab.~\ref{kitti_quantitative_result} and the qualitative results are shown in Fig.~\ref{kitti_qualitative_result}.

The quantitative results show that our model, FG-Depth, comprehensively exceeds all existing unsupervised methods that are even trained with stereo video(MS). Compared with \cite{Edge_Depth} which uses extra expensive semantic segmentation labels, our framework uses additional low-cost synthetic optical flow dataset and gets better performance. Compared with \cite{Guo_2018_ECCV} which also distills knowledge from optical flow network pretrianed on sceneflow dataset, FG-Depth performs better event at low resolution. Despite lack of the supervision from ground truth depth maps, the high-resolution performance of FG-Depth is even close to AdaBins~\cite{AdaBins}, a recent state-of-the-art supervised method, and FG-Depth has fewer parameters meanwhile. Besides, the qualitative results show that FG-Depth can produce sharper results even in some low-texture regions and on some thin structures. 

\begin{figure}[h]
    \centering
    \subfigure{
        \begin{minipage}[t]{0.19\linewidth}
            \centering
            {\scriptsize{Input}}
        \end{minipage}
        \begin{minipage}[t]{0.19\linewidth}
            \centering
            {\scriptsize{Monodepth2~\cite{Monodepth2}}}
        \end{minipage}
        \begin{minipage}[t]{0.19\linewidth}
            \centering
            {\scriptsize{EPCDepth~\cite{EPCDepth}}}
        \end{minipage}
        \begin{minipage}[t]{0.19\linewidth}
            \centering
            {\scriptsize{Ours(FG-Depth)}}
        \end{minipage}
        \begin{minipage}[t]{0.19\linewidth}
            \centering
            {\scriptsize{Ground truth}}
        \end{minipage}
    }\vspace{-3mm}
    \subfigure{
        \begin{minipage}[t]{0.19\linewidth}
            \centering
            \raisebox{-0.1cm}{\includegraphics[width=1.7cm]{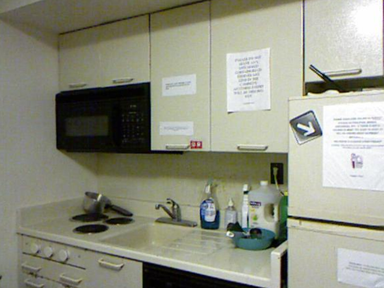}}
        \end{minipage}
        \begin{minipage}[t]{0.19\linewidth}
            \centering
            \raisebox{-0.1cm}{\includegraphics[width=1.7cm]{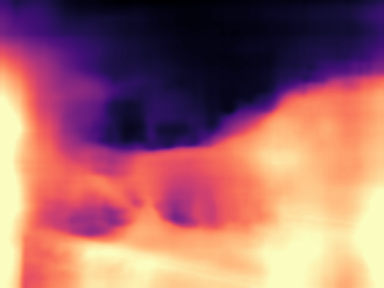}}
        \end{minipage}
        \begin{minipage}[t]{0.19\linewidth}
            \centering
            \raisebox{-0.1cm}{\includegraphics[width=1.7cm]{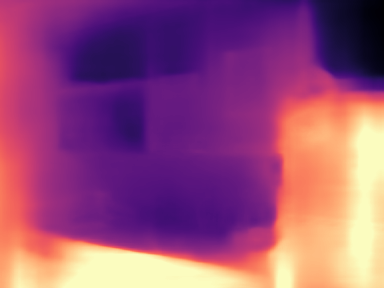}}
        \end{minipage}
        \begin{minipage}[t]{0.19\linewidth}
            \centering
            \raisebox{-0.1cm}{\includegraphics[width=1.7cm]{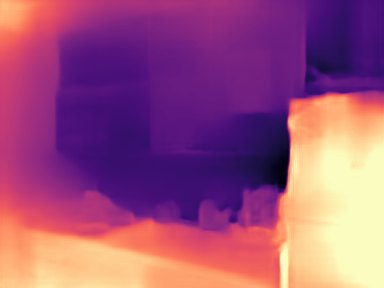}}
        \end{minipage}
        \begin{minipage}[t]{0.19\linewidth}
            \centering
            \raisebox{-0.1cm}{\includegraphics[width=1.7cm]{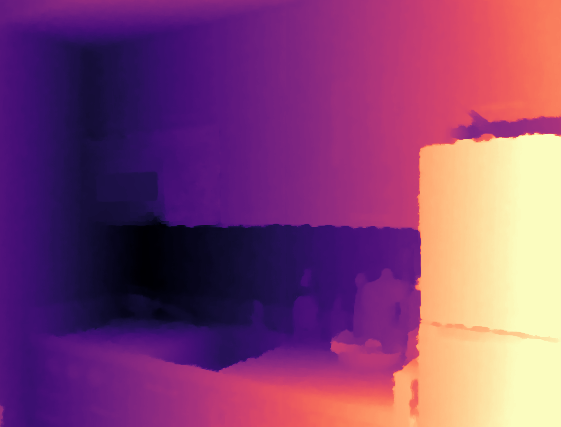}}
        \end{minipage}
    }\vspace{-3mm}
    \subfigure{
        \begin{minipage}[t]{0.19\linewidth}
            \centering
            \raisebox{-0.1cm}{\includegraphics[width=1.7cm]{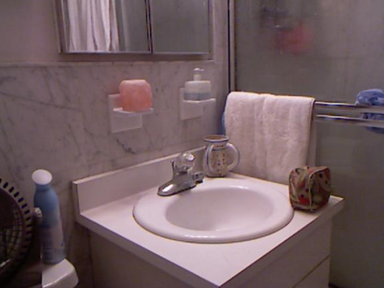}}
        \end{minipage}
        \begin{minipage}[t]{0.19\linewidth}
            \centering
            \raisebox{-0.1cm}{\includegraphics[width=1.7cm]{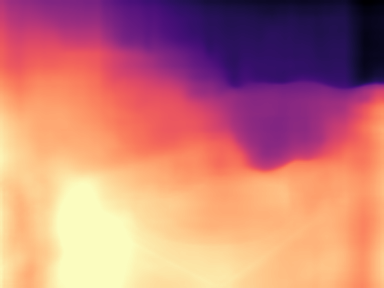}}
        \end{minipage}
        \begin{minipage}[t]{0.19\linewidth}
            \centering
            \raisebox{-0.1cm}{\includegraphics[width=1.7cm]{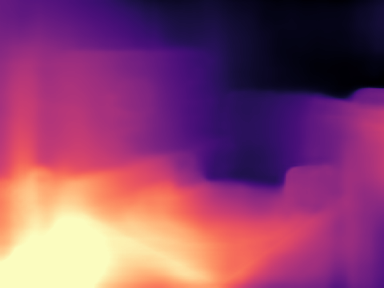}}
        \end{minipage}
        \begin{minipage}[t]{0.19\linewidth}
            \centering
            \raisebox{-0.1cm}{\includegraphics[width=1.7cm]{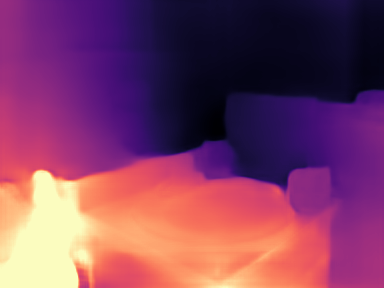}}
        \end{minipage}
        \begin{minipage}[t]{0.19\linewidth}
            \centering
            \raisebox{-0.1cm}{\includegraphics[width=1.7cm]{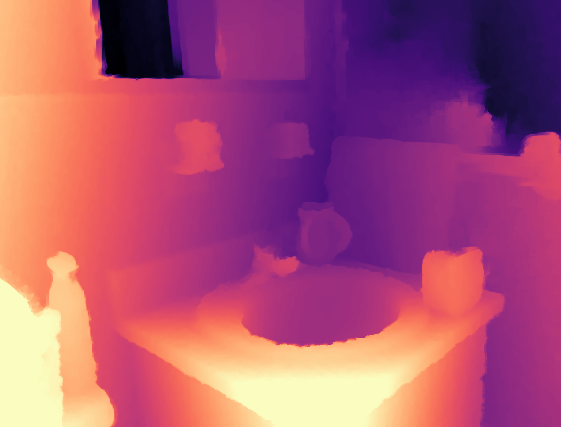}}
        \end{minipage}
    }\vspace{-3mm}
    \caption{\textbf{Qualitative results on the NYUV2 dataset.}}\label{nyuv2_generalization_qualitative_results}
\end{figure}

Then, we validate the performance on the NYU-Depth-v2 dataset using our model trained on the KITTI just as EPCDepth~\cite{Edge_Depth} did. The quantitative results in Tab.~\ref{nyuv2_generalization_quantitative_results} and the qualitative results in Fig.~\ref{nyuv2_generalization_qualitative_results} verify the strong generalization ability of our model.

\begin{table}[h]
    \caption{\textbf{Quantitative results on the NYU-Depth-v2 dataset.}}\label{nyuv2_generalization_quantitative_results}
    \centering
    \scriptsize
    \setlength\tabcolsep{3pt}
    \begin{tabular}{lccccccc}
        \hline
        \multicolumn{1}{l|}{Method} & \multicolumn{1}{c}{\cellcolor{red!30}Abs Rel} & \multicolumn{1}{c}{\cellcolor{red!30}Sq Rel} & \multicolumn{1}{c}{\cellcolor{red!30}RMSE} & \multicolumn{1}{c}{\cellcolor{red!30}RMSE$_{log}$} & \multicolumn{1}{c}{\cellcolor{blue!30}$\delta^{1}$} & \multicolumn{1}{c}{\cellcolor{blue!30}$\delta^{2}$} & \multicolumn{1}{c}{\cellcolor{blue!30}$\delta^{3}$}\\ 
        \hline
        \multicolumn{1}{l|}{Monodepth2~\cite{Monodepth2}}  & \multicolumn{1}{c}{0.362} & \multicolumn{1}{c}{0.718}  & \multicolumn{1}{c}{1.312}  & \multicolumn{1}{c}{0.384} & \multicolumn{1}{c}{0.477}  & \multicolumn{1}{c}{0.758}  & \multicolumn{1}{c}{0.898}           \\
        \multicolumn{1}{l|}{EPCDepth~\cite{EPCDepth}}  & \multicolumn{1}{c}{0.281} & \multicolumn{1}{c}{0.341}  & \multicolumn{1}{c}{0.912}  & \multicolumn{1}{c}{0.319} & \multicolumn{1}{c}{0.554}  & \multicolumn{1}{c}{0.833}  & \multicolumn{1}{c}{0.943}          \\
        \multicolumn{1}{l|}{Ours(FG-Depth)}  & \multicolumn{1}{c}{\textbf{0.269}} & \multicolumn{1}{c}{\textbf{0.318}}  & \multicolumn{1}{c}{\textbf{0.888}}  & \multicolumn{1}{c}{\textbf{0.312}} & \multicolumn{1}{c}{\textbf{0.560}}  & \multicolumn{1}{c}{\textbf{0.840}}  & \multicolumn{1}{c}{\textbf{0.947}} \\
        \hline
    \end{tabular}
\end{table}

\subsection{Ablation studies}
To better understand the effect of each component of our proposed model, we perform an ablation study and present the results in Tab.~\ref{ablation_result}. The results show that all our components can lead to significant performance when combined together. 

\begin{table}[h]
    \caption{\textbf{Ablation studies.} 
    $Base$ refers to the network architecture, $L_{p}$ refers to the photometric loss, $L_{fd}$ refers to the flow 
    distillation loss, $M_{p}$ refers to the auto mask proposed in \cite{Monodepth2} and $M_{f}$ refers to our 
    prior flow based mask.}\label{ablation_result}
    \centering
    \scriptsize
    \setlength\tabcolsep{3pt}
    \begin{tabular}{cccccccc}
        \hline
        \multicolumn{1}{c|}{Method} & \multicolumn{1}{c}{\cellcolor{red!30}Abs Rel} & \multicolumn{1}{c}{\cellcolor{red!30}Sq Rel} & \multicolumn{1}{c}{\cellcolor{red!30}RMSE} & \multicolumn{1}{c|}{\cellcolor{red!30}RMSE$_{log}$} & \multicolumn{1}{c}{\cellcolor{blue!30}$\delta^{1}$} & \multicolumn{1}{c}{\cellcolor{blue!30}$\delta^{2}$} & \multicolumn{1}{c}{\cellcolor{blue!30}$\delta^{3}$} \\ 
        \hline
        \multicolumn{1}{c|}{$Base$+$L_{P}$} & \multicolumn{1}{c}{0.106} & \multicolumn{1}{c}{1.300} & \multicolumn{1}{c}{5.850} & \multicolumn{1}{c|}{0.201} & \multicolumn{1}{c}{0.872} & \multicolumn{1}{c}{0.953} & \multicolumn{1}{c}{0.977}            \\
        \multicolumn{1}{c|}{$Base$+$L_{P}$+$M_{p}$} & \multicolumn{1}{c}{0.104} & \multicolumn{1}{c}{0.919} & \multicolumn{1}{c}{5.176} & \multicolumn{1}{c|}{0.202} & \multicolumn{1}{c}{0.873} & \multicolumn{1}{c}{0.953} & \multicolumn{1}{c}{0.976}            \\
        \multicolumn{1}{c|}{$Base$+$L_{P}$+$M_{f}$} & \multicolumn{1}{c}{0.100} & \multicolumn{1}{c}{0.730} & \multicolumn{1}{c}{4.499} & \multicolumn{1}{c|}{0.195} & \multicolumn{1}{c}{0.878} & \multicolumn{1}{c}{0.956} & \multicolumn{1}{c}{0.979}            \\
        \multicolumn{1}{c|}{$Base$+$L_{fd}$} & \multicolumn{1}{c}{0.099} & \multicolumn{1}{c}{1.102} & \multicolumn{1}{c}{5.230} & \multicolumn{1}{c|}{0.180} & \multicolumn{1}{c}{0.894} & \multicolumn{1}{c}{0.965} & \multicolumn{1}{c}{0.983} \\
        \multicolumn{1}{c|}{$Base$+$L_{fd}$+$M_{p}$} & \multicolumn{1}{c}{0.097} & \multicolumn{1}{c}{0.970} & \multicolumn{1}{c}{5.182} & \multicolumn{1}{c|}{0.180} & \multicolumn{1}{c}{0.896} & \multicolumn{1}{c}{0.965} & \multicolumn{1}{c}{0.983} \\ 
        \multicolumn{1}{c|}{$Base$+$L_{fd}$+$M_{f}$} & \multicolumn{1}{c}{\textbf{0.093}} & \multicolumn{1}{c}{\textbf{0.634}} & \multicolumn{1}{c}{\textbf{4.123}} & \multicolumn{1}{c|}{\textbf{0.174}} & \multicolumn{1}{c}{\textbf{0.900}} & \multicolumn{1}{c}{\textbf{0.967}} & \multicolumn{1}{c}{\textbf{0.984}} \\ 
    \hline
    \end{tabular}
\end{table}

\paragraph{Flow distillation loss} 
Although $L_{p}$ is common in previous self-supervised works, we show that it actually limits the capacity of models. Tab.~\ref{ablation_result} shows that in all cases, being trained with $L_{fd}$ can outperform those with $L_{p}$. 

\paragraph{Prior flow based mask}
Tab.~\ref{ablation_result} also shows that prior flow based mask $M_{f}$ significantly improves performance and its improvement is more significant than $M_{p}$ proposed in \cite{Monodepth2} even though $M_{f}$ dosen't remove occlusion for $L_{p}$ while $L_{fd}$ isn't disturbed by occlusion.

\paragraph{Loss function combinations} 
Tab.~\ref{loss_combinations} lists performance of different combinations of loss function. The results show that using $L_{dr}$ can already get impressive performance and combining $L_{dr}$ with $L_{fp}$ can get state-of-the-art performance which is consistent with the analysis in Sec.~\ref{lfd}. 

\paragraph{Pipeline} 
For a fair comparison with \cite{Guo_2018_ECCV}, we give results under different pipelines in Tab.~\ref{pipline}. The results on the first row and on the third row show that our networks have better performance even though at a smaller resolution. And the results in the last row show that our contributions can significantly improve the accuracy and even outperform \cite{Guo_2018_ECCV} finetuned with the supervised method. 

\begin{table}[h]
    \caption{\textbf{Ablation studies on loss function combinations.} 
    $L_{p}$ refers to the photometric loss, $L_{dr}$ refers to the depth regression loss and $L_{fp}$ refers to the flow-guided photometric loss.}\label{loss_combinations}
    \centering
    \scriptsize
    \setlength\tabcolsep{3pt}
    \begin{tabular}{cccccccc}
        \hline
        \multicolumn{1}{c|}{Loss} & \multicolumn{1}{c}{\cellcolor{red!30}Abs Rel} & \multicolumn{1}{c}{\cellcolor{red!30}Sq Rel} & \multicolumn{1}{c}{\cellcolor{red!30}RMSE} & \multicolumn{1}{c|}{\cellcolor{red!30}RMSE$_{log}$} & \multicolumn{1}{c}{\cellcolor{blue!30}$\delta^{1}$} & \multicolumn{1}{c}{\cellcolor{blue!30}$\delta^{2}$} & \multicolumn{1}{c}{\cellcolor{blue!30}$\delta^{3}$} \\ 
        \hline
        \multicolumn{1}{c|}{$L_{dr}$} & \multicolumn{1}{c}{0.094} & \multicolumn{1}{c}{0.643} & \multicolumn{1}{c}{4.139} & \multicolumn{1}{c|}{0.175} & \multicolumn{1}{c}{0.896} & \multicolumn{1}{c}{0.965} & \multicolumn{1}{c}{\textbf{0.985}}            \\
        \multicolumn{1}{c|}{$L_{fp}$} & \multicolumn{1}{c}{0.098} & \multicolumn{1}{c}{0.718} & \multicolumn{1}{c}{4.230} & \multicolumn{1}{c|}{0.177} & \multicolumn{1}{c}{0.892} & \multicolumn{1}{c}{0.966} & \multicolumn{1}{c}{0.984} \\
        \multicolumn{1}{c|}{$L_{dr}$+$L_{fp}$} & \multicolumn{1}{c}{\textbf{0.093}} & \multicolumn{1}{c}{\textbf{0.634}} & \multicolumn{1}{c}{\textbf{4.123}} & \multicolumn{1}{c|}{\textbf{0.174}} & \multicolumn{1}{c}{\textbf{0.900}} & \multicolumn{1}{c}{\textbf{0.967}} & \multicolumn{1}{c}{0.984} \\
        \hline
    \end{tabular}
\end{table}

\begin{table}[h]
    \caption{\textbf{Ablation studies on piplines.} $unsupFt$ and $supFt$ respectively refers to fituning the Flow-Net using unsupervised and supervised method on real data. $disp$ refers to using disparity to supervise the Depth-Net for all pixels with prediction of FLow-Net without fituning as the pipline on the first row did.}\label{pipline}
    \centering
    \tiny
    \setlength\tabcolsep{3pt}
    \begin{tabular}{ccccccccc}
        \hline
        \multicolumn{1}{c|}{pipline} & \multicolumn{1}{c|}{resolution} & \multicolumn{1}{c}{\cellcolor{red!30}Abs Rel} & \multicolumn{1}{c}{\cellcolor{red!30}Sq Rel} & \multicolumn{1}{c}{\cellcolor{red!30}RMSE} & \multicolumn{1}{c|}{\cellcolor{red!30}RMSE$_{log}$} & \multicolumn{1}{c}{\cellcolor{blue!30}$\delta^{1}$} & \multicolumn{1}{c}{\cellcolor{blue!30}$\delta^{2}$} & \multicolumn{1}{c}{\cellcolor{blue!30}$\delta^{3}$} \\ 
        \hline
        \multicolumn{1}{c|}{Guo~\cite{Guo_2018_ECCV} $w/o Ft$} & \multicolumn{1}{c|}{$384\times 1280$} & \multicolumn{1}{c}{0.109} & \multicolumn{1}{c}{0.822} & \multicolumn{1}{c}{4.656} & \multicolumn{1}{c|}{0.192} & \multicolumn{1}{c}{0.868} & \multicolumn{1}{c}{0.958} & \multicolumn{1}{c}{0.981}            \\
        \multicolumn{1}{c|}{Guo~\cite{Guo_2018_ECCV} $unsupFt$} & \multicolumn{1}{c|}{$384\times 1280$} & \multicolumn{1}{c}{0.099} & \multicolumn{1}{c}{0.745} & \multicolumn{1}{c}{4.424} & \multicolumn{1}{c|}{0.182} & \multicolumn{1}{c}{0.884} & \multicolumn{1}{c}{0.963} & \multicolumn{1}{c}{0.983} \\
        \multicolumn{1}{c|}{Guo~\cite{Guo_2018_ECCV} $supFt$} & \multicolumn{1}{c|}{$384\times 1280$} & \multicolumn{1}{c}{0.097} & \multicolumn{1}{c}{0.653} & \multicolumn{1}{c}{4.170} & \multicolumn{1}{c|}{0.170} & \multicolumn{1}{c}{0.889} & \multicolumn{1}{c}{\textbf{0.967}} & \multicolumn{1}{c}{\textbf{0.986}} \\
        \multicolumn{1}{c|}{ours($disp$)} & \multicolumn{1}{c|}{$192\times 640$} & \multicolumn{1}{c}{0.103} & \multicolumn{1}{c}{1.353} & \multicolumn{1}{c}{5.768} & \multicolumn{1}{c|}{0.185} & \multicolumn{1}{c}{0.891} & \multicolumn{1}{c}{0.964} & \multicolumn{1}{c}{0.982} \\
        \multicolumn{1}{c|}{ours($L_{dr}$+$L_{fp}$+$M_{f}$)} & \multicolumn{1}{c|}{$192\times 640$} & \multicolumn{1}{c}{\textbf{0.093}} & \multicolumn{1}{c}{\textbf{0.634}} & \multicolumn{1}{c}{\textbf{4.123}} & \multicolumn{1}{c|}{\textbf{0.174}} & \multicolumn{1}{c}{\textbf{0.900}} & \multicolumn{1}{c}{\textbf{0.967}} & \multicolumn{1}{c}{0.984} \\
        \hline
    \end{tabular}
\end{table}

\section{Conclusion}
In this paper, to break the bottleneck of unsupervised monocular depth estimation, noting that optical flow estimation models have strong generalization ability and the typical photometric loss is defective, we propose a flow distillation loss and a prior flow based mask to improve the performance of the unsupervised monocular depth estimator. And the experiments demonstrate that our model, FG-Depth, can achieve state-of-the-art performance on the KITTI dataset and NYU-Depth-v2 dataset. In future work, to further improve the performance, we will explore more methods to make full use of prior optical flow and try to apply our contributions to other categories that use monocular video(M) or stereo video(MS) as input.


\end{document}